\pdfoutput=1

\documentclass[11pt]{article}

\usepackage[preprint]{nlp4MusA}

\usepackage{parskip}

\usepackage{times}
\usepackage{amsmath}
\usepackage{latexsym}
\usepackage{multirow} 

\usepackage[T1]{fontenc}

\usepackage[utf8]{inputenc}

\usepackage{microtype}

\usepackage{inconsolata}

\usepackage{graphicx}

\title{MusiScene: Leveraging MU-LLaMA for Scene Imagination and Enhanced Video Background Music Generation}

\author{%
  Fathinah Izzati\thanks{These authors contributed equally to this work.}%
  \quad
  Xinyue Li\footnotemark[1]%
  \quad
  Yuxuan Wu%
  \quad
  Gus Xia\\
  Mohamed bin Zayed University of Artificial Intelligence\\
  \texttt{\{fathinah.izzati,xinyue.li,yuxuan.wu,gus.xia\}@mbzuai.ac.ae}
}

\begin{document}
\maketitle
\begin{abstract}

Humans can imagine various atmospheres and settings when listening to music, envisioning movie scenes that complement each piece. For example, slow, melancholic music might evoke scenes of heartbreak, while upbeat melodies suggest celebration. This paper explores whether a Music Language Model, e.g. MU-LLaMA, can perform a similar task, called Music Scene Imagination (MSI), which requires cross-modal information from video and music to train. To improve upon existing music captioning models which  focusing solely on musical elements, we introduce MusiScene, a music captioning model designed to imagine scenes that complement each music. In this paper, (1) we construct a large-scale video-audio caption dataset with 3,371 pairs, (2) we finetune Music Understanding LLaMA for the MSI task to create MusiScene, and (3) we conduct comprehensive evaluations and prove that our MusiScene is more capable of generating contextually relevant captions compared to MU-LLaMA. We leverage the generated MSI captions to enhance Video Background Music Generation (VBMG) from text. Code and datasets are available on GitHub\footnote{https://github.com/xinyueli2896/MusiScene.git} and huggingface\footnote{https://huggingface.co/datasets/tina2900/musi-scene}.

\end{abstract}

\section{Introduction}

Recently, \cite{herman} explored the use of large language models to generate background music for video. It uses the output of the video captioning models as input to the text-to-music model for generating music. Their project shares a similar goal with their predecessor in the same field, \cite{Di_2021}, who introduced a novel approach for generating video background music using a Controllable Music Transformer, which allows for the adjustment of musical attributes to align with video content. In our initial experiments, we found that music captioning enriches video descriptions with musical elements, enabling the generation of more nuanced soundtracks. However, we seek to improve further the generated music with new context-related captions and more efficient generation. We hypothesize that to generate background music from a video, it might be sufficient to use only music captions with their intrinsic scene, ambiance, atmosphere, and settings imagination abilities.

However, the problem with most existing music captioning models is that they mainly focus on categorizing the audio from musical aspects, such as key, tempo, and mood of the music, lacking the ability of scene imagination. To improve this ability, we present MusiScene, a music captioning model that addresses the additional challenge of Music Scene Imagination(MSI) task. We aim to explore whether a music LLM can imagine scenes given a music input and how this capability would affect the VBMG process. It is also potentially beneficial for downstream tasks such as enhancing music captions, tagging, and recommendation systems. Our contributions in this paper can be summarized as follows:
\begin{enumerate}
\item We finetune Music Understand LLaMA(MU-LLaMA) \cite{liu2023music}, the current state-of-the-art in general music question answering and captioning tasks, to answer scene-related questions such as "what kinds of video would this piece of music be suitable for?". This music scene imagination capability is a novel task in the domain of music understanding.

\item We curate a substantial cross-modal dataset containing videos with their background audio, video and music captions, along with a fusion of the two, for broader utilization in both video and music caption generation tasks. 

\item We conduct robust objective and subjective evaluations to assess the quality of music generated with MusiScene output captions. Objective evaluation results indicate in music generation task, music generated with MusiScene output captions is more plausible. On top of this, in subjective evaluation, music generated with MusiScene output captions get the average score of 74.2 when participants are asked to rate whether the background music suits the video, while music generated with pure music captions and video captions are rated at 73.5 and 61.4 respectively, reflecting higher coherence between MusiScene captions with the video.
\end{enumerate}

\begin{table*}[htbp]
\centering
\begin{tabular}{|l|p{13cm}|}
\hline
\textbf{Type} & \textbf{Caption} \\
\hline

Video Captions & A basketball game is being played in front of a crowd. \\
\hline
Music Captions & The music has a mood of suspense and tension, with a steady rhythm and a strong beat. It is a blend of different genres, including orchestral, electronic, and film music. \\
\hline
\multirow{2}{*}{MSI Captions(Ours)} & Prompt to Mixtral: Video Caption: "..", Music Caption: "..". What type of scene the music is suitable for? \\ 
\cline{2-2}
& \textbf{The music is suitable for a scene of sports competition, such as a crucial moment in a basketball game, where a high level of tension and excitement is being built up.} \\ 
\hline
\end{tabular}
\caption{A sample of finetuning data and its source}
\label{table:1}
\end{table*}

\section{Related Work}

\subsection{Video Background Music Generation}

Video Background Music Generation is the process of creating background music for videos to enhance their engagement and impact. This includes making social media content, advertisements, and movies more appealing. There has been effort in the field to generate background music for videos by using a Controllable Music Transformer \cite{Di_2021}. Herrmann-1 as one of the first to adopt an LLM-driven VBMG technique do so by extracting critical elements from a movie scene, including visual cues and dialogues \cite{herman}. It performs a comprehensive emotional analysis to generate contextually enriched text descriptions. These descriptions are subsequently transformed into detailed music conditions using GPT-4, guiding the model to produce a soundtrack that enhances the viewing experience.
\subsection{Music Captioning}
\label{sec:length}
The Music Understanding LLaMA (MU-LLaMA) \cite{liu2023music} model marks a significant leap in text-to-music generation, tackling the challenge of limited music datasets annotated with natural language. By combining music-related question answering and caption generation, leveraging the MERT model \cite{mert} for audio feature extraction, MU-LLaMA introduces an efficient and scalable approach to music captioning. Its architecture incorporates a pre-trained MERT Encoder and a Music Understanding Adapter, enabling it to grasp complex musical contexts.

Despite the high-quality captions generated by existing models, the captions broadly include monotonous and vague descriptions such as "The music is upbeat and energetic" or "The music is smooth and is in the type of blue", which might not be helpful for some music generation tasks that require more depth and accuracy, such as video background music generation. 

Another contemporary music captioning model, LP-MusicCaps \cite{doh2023lpmusiccaps}, demonstrates the ability to generate captions from music tags at a tag-to-caption level. However, it still relies on predicting metadata from the music before generating captions, thereby limiting its applicability to scenarios where direct audio-to-caption generation is needed.

\section{Method}
\subsection{Dataset Creation}
To finetune MU-LLaMA for the scene imagination task, we need training data of background music and the expected output shown in Table 1 (MSI). Since there is not a readily available dataset that serves this purpose, we constructed our own dataset: Video-Audio CAptions Dataset (VACAD). The foundation of our dataset is Audioset \cite{7952261}, an extensive dataset created by Google Research, consisting of an expansive collection of over 2 million labelled 10-second sound clips drawn from YouTube videos. These clips are labelled with 632 audio event classes, including human sounds, animal sounds, musical instruments, and ambient noises. We chose 3371 video clips from its test set with the class label 'Music'. As shown in Table 1, we captioned the video with SwinBERT \cite{lin2022swinbert} and the background music with MU-LLaMA. To combine cross-modal captions and to generate ground truth for MSI, we use Mixtral of Experts \cite{jiang2024mixtral} LLM with different prompts each (see Appendix A).

\subsection{Finetune MU-LLaMA}
We use the pre-trained MU-LLaMA model and finetune it for music scene imagination. The model for MU-LLaMA starts with the MERT Encoder. Then, the system processes raw music inputs to generate meaningful encoded features. These are then passed to a music understanding adapter, which integrates a Conv1D feature aggregator and several dens layers to refine and project these features. During finetune, we froze the LLaMA model \cite{llama} and finetuned the parameters in the music understanding adapter. The process of finetune is shown in Figure 1.

\begin{figure*}[htbp]
    \centering
    \includegraphics[scale = 0.32]{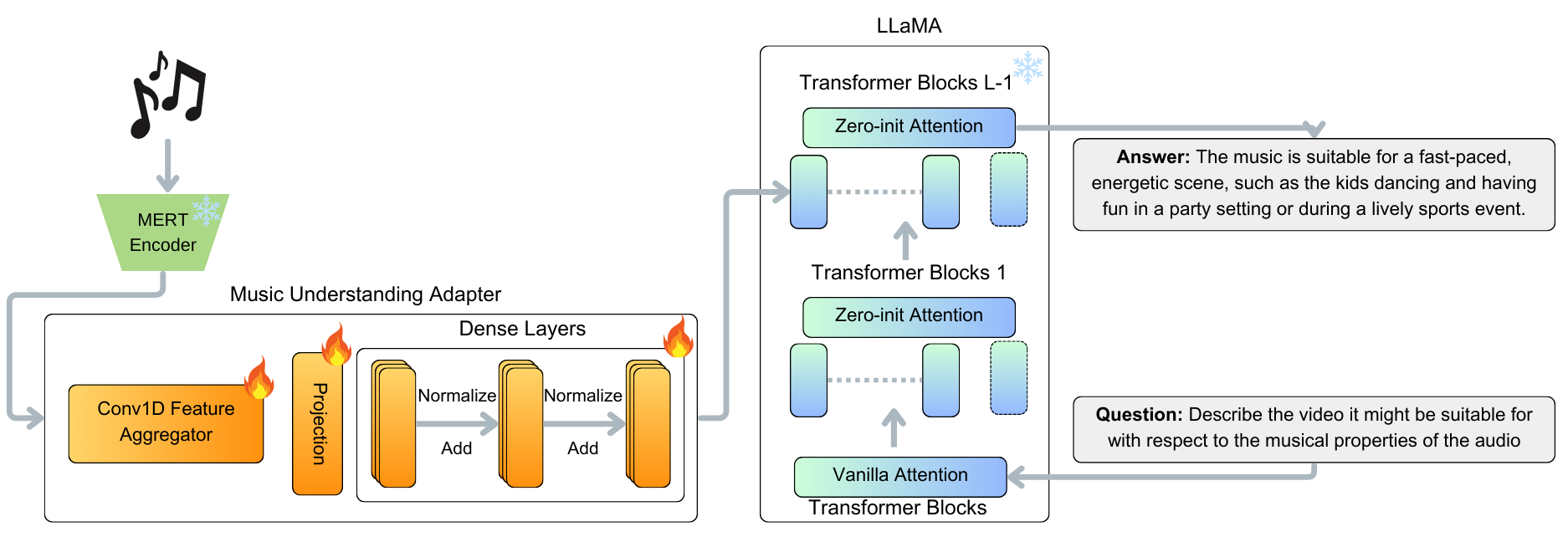}
    \caption{Model Architecture of MusiScene: Finetune MU-LLaMA for MSI task}
    \label{fig:enter-label}
\end{figure*}

\section{Experimental Setup}
\subsection{Training Setup}
For finetune, we use LLaMA 7B-chat because it has shown to balance the computational efficiency and performance to the greatest extent. Moreover, the architecture and training data of the 7B-chat model are better suited for capturing the nuances of music-related descriptions. We fintune on 80\% of the dataset and used 20\% of dataset as a test set. The sample data and its prompt is shown in MSI part of Table 1. The finetune runs through 20 epochs and takes approximately 5 hours to train on 4 A100 GPUs.

\subsection{Evaluation Setup}
We evaluated MusiScene using two different approaches: 1) Music Scene Imagination (MSI) task, and 2) Text to Music Generation using MusicGen. In the first approach, we compare MusiScene to the base model, MU-LLaMA, as baseline, and in the second approach, we compare MusiScene to SwinBERT video caption output and MU-LLaMA music caption output, and finally, a fusion of the two. Our goal is to prove that MusiScene as a music q\&a and captioning model can do scene imagination from given music. Further, we will prove that MSI captions can be used to generate music with higher context relevance than pure music and video captions.

For the second approach, we fed music scene imagination (MSI) captions into MusicGen \cite{musicgen} and evaluated the fit and quality of the generated music both objectively and subjectively. We compared the results with those generated from video captions, music captions, and fusion.

For objective metrics, we report the Fréchet Audio Distance (FAD) \cite{fad} and KL Divergence \cite{audiogen}, as applied by \cite{musicgen}. For FAD, we used the implementation from Google Research with the TensorFlow VGGish model \footnote{https://github.com/google-research/google-research/tree/master/frechet\_audio\_distance}.This model creates embeddings of audio files to extract features from audio data. A low FAD score indicates that the generated audio is plausible. To compute the KL Divergence, we used implementation from Meta Audiocraft with the PASST classifier \footnote{https://github.com/facebookresearch/audiocraft/}. 
This metric calculate the probability distributions of labels for both the target and predicted audio.

For subjective evaluation, we followed a similar setup to \cite{yang}. We collected responses from two forms, with 68 and 50 respondents respectively. The respondents were university students, ranging from undergraduates to postgraduates, with no requirement for a musical background. In the first form, we presented 15 random videos, each containing three different tracks generated from video captions, music captions, and MSI captions. In the second form, we presented 8 random videos, each containing two different tracks generated from fusion captions and MSI captions. 

\section{Results}



\subsection{Objective Evaluation}
\paragraph{MSI} We evaluate the power of music scene imagination of MusiScene in comparison to the original MU-LLaMA. The evaluation results shown in the table below indicate a significant difference in their performance across four natural language processing (NLP) metrics, BLEU (B-U), METEOR (M-R), ROUGEL (R-L) and BERT-Score (BERT-S). The BLEU score is a weighted average of $\text{BLEU}_{1-4}$.

\begin{table}[htbp]
\centering
\begin{tabular}{|l|c|c|c|c|}
\hline
\textbf{Model} & \textbf{B-U} & \textbf{M-R} & \textbf{R-L} & \textbf{B-S} \\
\hline
MU-LLaMA & 0.190 & 0.207 & 0.219 & 0.863 \\
\hline
\textbf{MusiScene} & \textbf{0.421} & \textbf{0.403} & \textbf{0.404} & \textbf{0.901} \\
\hline
\end{tabular}
\caption{Performance comparison of two model checkpoints.}
\label{table:2}
\end{table}

These metrics measure how close the generated captions are to the referenced captions created in Section 3. The metrics shown in Table 2 
suggests that MusiScene produces captions that are more context-related and capable of scene imagination than MU-LLaMA. 

\paragraph{Text2Music Gen}  
For the objective evaluation of the text-to-music generation task, our results indicate that MSI captions generate relatively plausible music under both metrics. However, video captions and fusion captions can also produce plausible music. 

\begin{table}[htbp]
\centering
\begin{tabular}{|l|c|c|c|c|}
\hline
\textbf{Text Input} & \textbf{FAD} & \textbf{KL Div.} \\
\hline
\textbf{MSI (Ours)} & \textbf{5.78} & \textbf{2.10}  \\
Video Caps & \textbf{4.46} & 2.13 \\
Music Caps & 6.43 & 2.12 \\
Fusion & 6.62 & \textbf{2.05} \\
\hline
\end{tabular}
\caption{Objective Evaluation Result on Text2Music Downstream Task}
\label{table:3}
\end{table}

Although a lower FAD indicates that the generated audio is closer to the actual audio in terms of distribution, this does not necessarily imply better quality. This is the case for video captions where it achieves the highest FAD score yet scored the lowest in KL Divergence and subjective evaluation.

\subsection{Subjective Evaluation}

Table 4 below shows the average scores of the 15 x 3 video-music evaluations by 68 respondents. Table 5 below shows the average scores of the 8 x 2 video-music evaluations by 50 respondents.

\begin{table}[htbp]
\centering
\begin{tabular}{|l|c|c|c|c|}
\hline
 \textbf{MSI} & \textbf{Music Caps} & \textbf{Video Caps}  \\
\hline
74.2 & 73.5 & 61.4 \\
\hline
\end{tabular}
\caption{Subjective Evaluation Result on Text2Music Downstream Task}
\label{table:4}
\end{table}
\begin{table}[htbp]
\centering
\begin{tabular}{|l|c|c|c|c|}
\hline
\textbf{MSI} & \textbf{Fusion} \\
\hline
 78.4 & 76.6 \\
\hline
\end{tabular}
\caption{Subjective Evaluation Result on Text2Music Downstream Task}
\label{table:5}
\end{table}
The results indicate that MSI outperforms all other captioning strategies for text-to-music generation. Surprisingly, despite the very verbose captions from Fusion, MSI achieves a higher score. MSI outperforms both video-only and music-only captions, supporting our hypothesis that cross-modal information can improve text-to-music performance.
\section{Conclusion}
In this paper, we introduced MusiScene, a novel music captioning model designed to bridge the gap between audio and video through the Music Scene Imagination task. By fine-tuning the MU-LLaMA model with the extensive cross-modal dataset to answer scene-related questions, we enabled it to imagine appropriate video contexts for given musical pieces. The preferable evaluation results showed that MusiScene-generated captions led to more coherent and context-related music outputs compared to those based solely on musical or video features over other models. Our research demonstrated that integrating sophisticated music captioning modules into video soundtrack generation can enhance the quality of the generated music. Additionally, we found that using only the MSI captions from MusiScene is sufficient to capture scene-related information without seeing the actual video, and still achieving a contextually relevant video soundtrack in Video Background Music Generation.

\bibliography{nlp4MusA}

\begin{thebibliography}{13}
\providecommand{\natexlab}[1]{#1}

\bibitem[{Copet et~al.(2023)Copet, Kreuk, Gat, Remez, Kant, Synnaeve, Adi, and Défossez}]{musicgen}
Jade Copet, Felix Kreuk, Itai Gat, Tal Remez, David Kant, Gabriel Synnaeve, Yossi Adi, and Alexandre Défossez. 2023.
\newblock \href {https://arxiv.org/abs/2306.05284} {Simple and controllable music generation}.

\bibitem[{Di et~al.(2021)Di, Jiang, Liu, Wang, Zhu, He, Liu, and Yan}]{Di_2021}
Shangzhe Di, Zeren Jiang, Si~Liu, Zhaokai Wang, Leyan Zhu, Zexin He, Hongming Liu, and Shuicheng Yan. 2021.
\newblock \href {https://doi.org/10.1145/3474085.3475195} {Video background music generation with controllable music transformer}.
\newblock In \emph{Proceedings of the 29th ACM International Conference on Multimedia}, MM ’21. ACM.

\bibitem[{Doh et~al.(2023)Doh, Choi, Lee, and Nam}]{doh2023lpmusiccaps}
SeungHeon Doh, Keunwoo Choi, Jongpil Lee, and Juhan Nam. 2023.
\newblock \href {https://arxiv.org/abs/2307.16372} {Lp-musiccaps: Llm-based pseudo music captioning}.
\newblock \emph{Preprint}, arXiv:2307.16372.

\bibitem[{Gemmeke et~al.(2017)Gemmeke, Ellis, Freedman, Jansen, Lawrence, Moore, Plakal, and Ritter}]{7952261}
Jort~F. Gemmeke, Daniel P.~W. Ellis, Dylan Freedman, Aren Jansen, Wade Lawrence, R.~Channing Moore, Manoj Plakal, and Marvin Ritter. 2017.
\newblock \href {https://doi.org/10.1109/ICASSP.2017.7952261} {Audio set: An ontology and human-labeled dataset for audio events}.
\newblock In \emph{2017 IEEE International Conference on Acoustics, Speech and Signal Processing (ICASSP)}, pages 776--780.

\bibitem[{Haseeb et~al.(2024)Haseeb, Hammoudeh, and Xia}]{herman}
Muhammad~Taimoor Haseeb, Ahmad Hammoudeh, and Gus Xia. 2024.
\newblock \href {https://audiomatic-research.github.io/herrmann-1/} {Gpt-4 driven cinematic music generation through text processing}.

\bibitem[{{Hugo Touvron et al.}(2023)}]{llama}
{Hugo Touvron et al.} 2023.
\newblock \href {https://arxiv.org/abs/2307.09288} {Llama 2: Open foundation and fine-tuned chat models}.
\newblock \emph{Preprint}, arXiv:2307.09288.

\bibitem[{Jiang et~al.(2024)Jiang, Sablayrolles, Roux, Mensch, Savary, Bamford, Chaplot, de~las Casas, Hanna, Bressand, Lengyel, Bour, Lample, Lavaud, Saulnier, Lachaux, Stock, Subramanian, Yang, Antoniak, Scao, Gervet, Lavril, Wang, Lacroix, and Sayed}]{jiang2024mixtral}
Albert~Q. Jiang, Alexandre Sablayrolles, Antoine Roux, Arthur Mensch, Blanche Savary, Chris Bamford, Devendra~Singh Chaplot, Diego de~las Casas, Emma~Bou Hanna, Florian Bressand, Gianna Lengyel, Guillaume Bour, Guillaume Lample, Lélio~Renard Lavaud, Lucile Saulnier, Marie-Anne Lachaux, Pierre Stock, Sandeep Subramanian, Sophia Yang, Szymon Antoniak, Teven~Le Scao, Théophile Gervet, Thibaut Lavril, Thomas Wang, Timothée Lacroix, and William~El Sayed. 2024.
\newblock \href {https://arxiv.org/abs/2401.04088} {Mixtral of experts}.
\newblock \emph{Preprint}, arXiv:2401.04088.

\bibitem[{Kilgour et~al.(2018)Kilgour, Zuluaga, Roblek, and Sharifi}]{fad}
Kevin Kilgour, Mauricio Zuluaga, Dominik Roblek, and Matthew Sharifi. 2018.
\newblock \href {https://arxiv.org/abs/1812.08466} {Fréchet audio distance: A metric for evaluating music enhancement algorithms}.

\bibitem[{Kreuk et~al.(2022)Kreuk, Synnaeve, Polyak, Singer, Défossez, Copet, Parikh, Taigman, and Adi}]{audiogen}
Felix Kreuk, Gabriel Synnaeve, Adam Polyak, Uriel Singer, Alexandre Défossez, Jade Copet, Devi Parikh, Yaniv Taigman, and Yossi Adi. 2022.
\newblock \href {https://arxiv.org/abs/2209.15352} {Audiogen: Textually guided audio generation}.

\bibitem[{Li et~al.(2023)Li, Yuan, Zhang, Ma, Chen, Yin, Xiao, Lin, Ragni, Benetos, Gyenge, Dannenberg, Liu, Chen, Xia, Shi, Huang, Wang, Guo, and Fu}]{mert}
Yizhi Li, Ruibin Yuan, Ge~Zhang, Yinghao Ma, Xingran Chen, Hanzhi Yin, Chenghao Xiao, Chenghua Lin, Anton Ragni, Emmanouil Benetos, Norbert Gyenge, Roger Dannenberg, Ruibo Liu, Wenhu Chen, Gus Xia, Yemin Shi, Wenhao Huang, Zili Wang, Yike Guo, and Jie Fu. 2023.
\newblock \href {https://arxiv.org/abs/2306.00107} {Mert: Acoustic music understanding model with large-scale self-supervised training}.
\newblock \emph{Preprint}, arXiv:2306.00107.

\bibitem[{Lin et~al.(2022)Lin, Li, Lin, Ahmed, Gan, Liu, Lu, and Wang}]{lin2022swinbert}
Kevin Lin, Linjie Li, Chung-Ching Lin, Faisal Ahmed, Zhe Gan, Zicheng Liu, Yumao Lu, and Lijuan Wang. 2022.
\newblock \href {https://arxiv.org/abs/2111.13196} {Swinbert: End-to-end transformers with sparse attention for video captioning}.
\newblock \emph{Preprint}, arXiv:2111.13196.

\bibitem[{Liu et~al.(2023)Liu, Hussain, Sun, and Shan}]{liu2023music}
Shansong Liu, Atin~Sakkeer Hussain, Chenshuo Sun, and Ying Shan. 2023.
\newblock \href {https://arxiv.org/abs/2308.11276} {Music understanding llama: Advancing text-to-music generation with question answering and captioning}.
\newblock \emph{Preprint}, arXiv:2308.11276.

\bibitem[{Yang et~al.(2022)Yang, Yu, Wang, Wang, Weng, Zou, and Yu}]{yang}
Dongchao Yang, Jianwei Yu, Helin Wang, Wen Wang, Chao Weng, Yuexian Zou, and Dong Yu. 2022.
\newblock \href {https://arxiv.org/abs/2207.09983} {Diffsound: Discrete diffusion model for text-to-sound generation}.

\end{thebibliography}
\appendix

\section{Dataset}
\label{sec:appendix}
\begin{table}[htbp]
\centering
\begin{tabular}{|l|p{5cm}|}
\hline
\textbf{Method} & \textbf{Caption} \\
\hline

VideoCaps & A basketball game is being played in front of a crowd. \\
\hline
MusicCaps & The music has a mood of suspense and tension, with a steady rhythm and a strong beat. It is a blend of different genres, including orchestral, electronic, and film music. \\
\hline
\multirow{2}{*}{Fusion} & Prompt to Mixtral: Video Caption: "..", Music Caption: "..". Describe the music from both video and music captions. \\ 
\cline{2-2}
& The music described in the captions for the basketball game video and the music selection appears to be a dramatic and tense composition ... Overall, the music described is a dynamic and impactful composition that is well-suited to the fast-paced and thrilling atmosphere of a basketball game. \\
\hline
\multirow{2}{*} {MSI} & Prompt to Mixtral: Video Caption: "..", Music Caption: "..". What type of scene the music is suitable for? \\ 
\cline{2-2}
& The music is suitable for a scene of sports competition, such as a crucial moment in a basketball game, where a high level of tension and excitement is being built up. \\ 
\hline
\end{tabular}

\label{table:6}
\caption{A sample of fine-tuning data and its source}
\end{table}

\end{document}